\pdfoutput=1

\documentclass[11pt]{article}

\usepackage[]{acl}

\usepackage{listings}
\usepackage{times}
\usepackage{latexsym}
\usepackage{colortbl}
\usepackage{xcolor}
\usepackage{tabularx}
\usepackage{multirow}
\usepackage{amsmath}
\usepackage{graphicx}
\usepackage{amsfonts}
\usepackage{makecell}
\usepackage{xcolor}
\usepackage{tcolorbox}
\usepackage{booktabs}

\usepackage[T1]{fontenc}

\usepackage[utf8]{inputenc}

\usepackage{microtype}

\usepackage{inconsolata}

\title{Persuasiveness of Generated Free-Text Rationales in Subjective Decisions: A Case Study on Pairwise Argument Ranking}

\author{
  Mohamed Elaraby \textsuperscript{1} \quad Diane Litman \textsuperscript{1} \quad \textbf{Xiang Lorraine Li\textsuperscript{1}}   
   \quad \textbf{Ahmed Magooda \textsuperscript{2}}  \\
  \textsuperscript{1} University of Pittsburgh, Pittsburgh, PA, USA \\
  \textsuperscript{2} Microsoft, Redmond, WA, USA \\
  {\tt \{mse30,dlitman,xianglli\}@pitt.edu} \\ 
  {\tt ahmedmagooda@microsoft.com}
}

\begin{document}
\maketitle
\begin{abstract}

Generating free-text rationales
is among the emergent capabilities of Large Language Models (LLMs). These rationales have been found to enhance LLM performance across various NLP tasks. Recently, there has been growing interest in using these rationales to provide insights for various important downstream tasks. In this paper, we analyze generated free-text rationales in tasks with subjective answers, emphasizing the importance of rationalization in such scenarios. We focus on \textit{pairwise argument ranking}, a highly subjective task with significant potential for real-world applications, such as debate assistance. We evaluate the \textit{persuasiveness} of rationales generated by nine LLMs to support their subjective choices. Our findings suggest that open-source LLMs, particularly Llama2-70B-chat, are capable of providing highly persuasive rationalizations, surpassing even GPT models. Additionally, our experiments show that rationale persuasiveness can be improved by controlling its parameters through prompting or through self-refinement.
\end{abstract}

\section{Introduction}

Large Language Models (LLMs) \cite{brown2020language, chowdhery2022palm, scao2022bloom, touvron2023llama} have demonstrated a strong ability to generate \textit{free-text rationales} to explain and support their decisions in plain natural language, which adds an essential layer of transparency and interpretability to their outputs. Recently, there has been a growing interest in utilizing these rationales to enhance the usability and reliability of LLM-based applications, thereby reducing the risks posed by LLMs in decision-making processes \cite{bender2021dangers}.

\begin{figure}[!ht]
    \centering
    \includegraphics[height=0.65\textwidth]{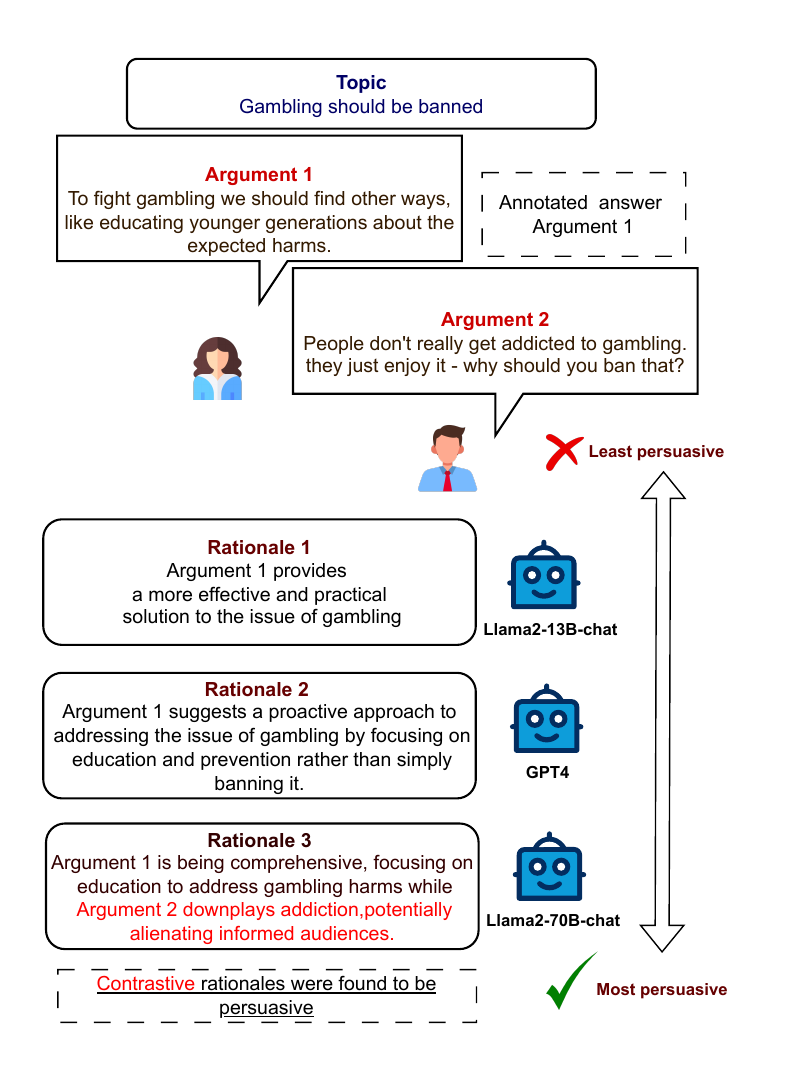}
    
   \caption{\centering Given two arguments with the same stance on a topic, the model selects the higher quality argument and generates a convincing rationale. We analyze the \textit{persuasiveness} of these rationales.}
   
   \label{fig:pairwise_task}
\end{figure}



Existing research on evaluating and analyzing free-text rationales has primarily focused on tasks where there is an expected factual ground truth answer that the model should achieve even without further explanation. Most of this work has focused on assessing the plausibility \cite{wiegreffe2021teach, marasovic2022few} and faithfulness \cite{wiegreffe2021measuring} of these rationales to produce accurate answers. Recently, studies have been introduced to also analyze 
rationales for their utility in learning new concepts \cite{joshi2023machine} and truth verification \cite{si2023large}. 

In this work, we analyze free-text rationales in subjective tasks where annotations, despite agreement, remain subjective. We focus on \textit{rationale persuasiveness} to understand how different LLMs convincingly justify their choices. Specifically, we examine rationales in \textit{pairwise argument ranking} \cite{gretz2020large, toledo2019automatic}, a task with inherent subjectivity and significant potential for applications like debate assistance tools \cite{wachsmuth2024argument}. In this task, the model recommends one argument from a pair on a controversial topic. We believe that adding persuasive rationales to argument recommendations can enhance their utility in downstream applications. Figure \ref{fig:pairwise_task} shows examples of rationales generated by various models. While these models agree on the pairwise ranking, their rationales reveal different levels of persuasiveness in supporting Argument 1.

We provide a comprehensive analysis of the persuasive nature of free-text rationales by addressing the following research questions (RQs):  \textbf{RQ1:} \textit{ How do different LLMs compare in generating persuasive rationales?} \textbf{RQ2:} \textit{Can we automatically detect the more persuasive rationales?} \textbf{RQ3:} \textit{Which characteristics of a rationale contribute to its persuasiveness?} \textbf{RQ4:} \textit{Can we control the persuasiveness of generated rationales?}
To address these questions, we: $(1)$ Prompt $9$ different LLMs to perform zero-shot pairwise ranking and provide rationales for their choices.
$(2)$ Use manually annotated rationales to evaluate automatic persuasiveness detection methods, specifically GPT4 \cite{openai2023gpt4}, for ranking rationale persuasiveness, enabling large-scale analysis.
$(3)$ Conduct a human evaluation study to rank the persuasiveness of generated rationales and examine the influence of the rationale's content.
 $(4)$ Experiment with enhancing rationale persuasion by prompting the model with key aspects for persuasion learned from prior steps and explore automatic self-improvement techniques to assess if the model can improve its persuasiveness.

Our findings can be summarized in four key points:
$(1)$ Open-source LLMs, particularly Llama2-70B-chat, excelled in generating persuasive rationales, even outperforming GPT4.
$(2)$ GPT4 closely matched human rankings of the persuasiveness of the rationales, although a perfect agreement was unattainable due to the inherent subjectivity of the task.
$(3)$ Contrastive rationales, which justify why the alternative argument was not chosen, emerged as the most influential factor in persuasiveness.
$(4)$ Prompting the model with persuasiveness factors can enhance the persuasiveness of the generated rationales.

\section{Related Work}


\paragraph{Argument Quality Ranking} 
Argument quality ranking is a key task in argument quality estimation, which can be approached in two main settings: \textit{(1) pointwise ranking,} where arguments are individually assessed based on a quality score like interpretability \cite{swanson2015argument}, human quality annotations \cite{toledo2019automatic, gretz2020large}; and \textit{(2) pairwise ranking}, where the quality of the arguments is estimated in comparison to each other, using factors such as persuasiveness \cite{habernal2016argument, simpson2018finding} or aggregated preferences \cite{toledo2019automatic}. \textit{Our work adopts the pairwise ranking framework in a zero-shot setting}.

\paragraph{LLMs for Argument Quality Ranking}
Despite their strong performance in various tasks, \citet{wang2023contextual} demonstrated that LLMs, particularly the GPT-3.5-turbo, struggle to match supervised models in point-wise and pair-wise ranking tasks, even in few-shot settings. Instead of relying solely on existing benchmarks, \citet{mirzakhmedova2024large} showed that LLMs, especially PALM2 and GPT-3, are effective in annotating argument quality, particularly when combined with human annotations. Recently, \citet{wachsmuth2024argument} suggested that LLMs could open new directions in argument quality research, such as fact-checking and argument optimization.
\textit{In this work, we analyze the persuasiveness of rationales generated by different LLMs, proposing that LLMs can enhance argument quality-based applications by providing users with persuasive explanations to support their decisions.}

\begin{table*}[ht]
\small
\centering
\resizebox{\textwidth}{!}{
\begin{tabular}{l|c|c|c|cc}
\toprule
\multirow{2}{*}{\textbf{Dataset}} & \multirow{2}{*}{\textbf{\# Argument Pairs (Unfiltered)}} & \multirow{2}{*}{\textbf{\# Argument Pairs (Filtered)}} & \multirow{2}{*}{\textbf{\# Rationales}} & \multicolumn{2}{c}{\textbf{\# Rationale Pairs for Persuasion}} \\ 
 &  &  &  & \textbf{Annotated} & \textbf{Full} \\ \hline
\textbf{IBM-9k}           & 400                                     & 30                                    & 270                    & 204               & 1080 \\ 
\textbf{IBM-30k}          & 1534                                    & 144                                   & 1296                   & -                 & 5184 \\ \bottomrule
\end{tabular}}
\caption{\centering 
Summary of datasets for evaluating free-text rationales. Unfiltered is the total argument pairs sampled, Filtered is the subset with unanimous LLM agreement, and Annotated is the subset used for human evaluation.}
\label{tab:dataset_summary}
\end{table*}

\paragraph{Evaluating Free-Text Rationalization}
Evaluating free-text rationales has primarily focused on their ability to aid models in reaching correct answers. Metrics such as accuracy differences between predictions with and without rationales \cite{hase2020leakage, wiegreffe2021measuring} and information-theoretic measures \cite{chen2023rev} assess how rationale content supports model performance. \citet{wiegreffe2021teach} established criteria for evaluating rationales, including surface form for validity and grammatical correctness, support for association between the rationale and the label, and \textit{contrast} with alternative labels. Building on this, \citet{joshi2023machine} introduced \textit{novelty}, measuring the extent of new information provided by the rationale, enhancing its utility in human-ai collaboration tasks. In the context of persuasiveness, \citet{ajwani2024llm} found that LLMs can convincingly support incorrect predictions in the NLI task.
\textit{Given our study's focus is close to rationale utility, we adopt the dimensions introduced by \citet{joshi2023machine} to evaluate our rationale content.
We focus on persuasiveness for subjective tasks like pairwise argument ranking. We also included a large number of models and evaluation measures}.

\paragraph{Persuasiveness in LLMs}
Prior research on the persuasiveness of LLMs has compared generated arguments with those written by humans. \citet{bai2023artificial} conducted a randomized control trial showing that GPT-3 can write persuasive political arguments comparable to human ones. Similarly, \citet{palmer2023large} found that GPT-3's texts on controversial topics were as persuasive as those written by crowdsource workers. \citet{salvi2024conversational} demonstrated that personalization enhances GPT4's persuasiveness in conversations. \citet{rescala2024can} also showed that GPT4 can detect persuasiveness in debates as effectively as crowdsource workers.
\textit{However, most of this research has focused on large commercial LLMs and analyzing the arguments themselves. We shift the focus to the persuasiveness of rationales.
Additionally, we include a broader range of LLMs for a more comprehensive analysis.}

\section{Experimental Settings}


\subsection{Datasets}

To assemble the free-text rationales evaluation set, we used argument pairs from two datasets: \textit{IBM-ArgQ-9.1kPairs} (IBM-9k) \cite{toledo2019automatic} and \textit{IBM-30k-rank} (IBM-30k) \cite{gretz2020large}. The IBM-9k dataset contains pairs of arguments either supporting or opposing a topic, with annotations for the higher-quality argument. The IBM-30k dataset includes individual arguments annotated with quality scores ranging from $0$ to $1$.

From the IBM-9k dataset, we randomly selected $400$ argument pairs from the test set, evenly distributed across $20$ topics. This set was used for manual analysis and evaluation due to its quality control measures, which ensure that argument pairs advocate the same stance, are of high quality, and have comparable lengths to avoid length bias \cite{potash-etal-2017-length}. These pairs were used to prompt the LLMs for argument predictions and supporting rationales. We filtered out pairs where any LLM failed to predict the annotated winning argument, focusing on pairs with unanimous agreement to ensure a fair comparison between the generated rationales. This left us with $30$ argument pairs \footnote{Appendix \ref{app:agreed_upon} shows that considering agreement among all models leads to a significant reduction in the number of argument pairs}, each with rationales generated by $9$ models, totaling $270$ rationales. Comparing these rationales for persuasiveness resulted in $1080$ rationale pairs for evaluation.

For the IBM-30k dataset, we created a pairwise ranking set by sampling arguments that (1) have a similar stance, (2) vary in length by a maximum of 20\% to avoid bias, (3) each appear at most once to diversify the comparison set while reducing computation cost of prompting, and (4) have different quality scores, allowing us to assess the influence of the quality differences on the persuasiveness. This resulted in $1534$ pairs. We followed a similar prompting and filtering technique used for the IBM-9k dataset, which left us with $144$ unanimously agreed upon pairs, totaling $144*9=1296$ rationales. Comparing these rationales for persuasiveness resulted in $5184$ persuasion pairs. This dataset acts as an extended test set to assess whether our findings on the IBM-9k dataset will generalize to other topics and arguments. Table \ref{tab:dataset_summary} \footnote{The annotated data and code are to be released with the final version.} shows the  statistics of the datasets included in our work.

\begin{figure*}[ht]
\begin{center}
\includegraphics[scale=.62]{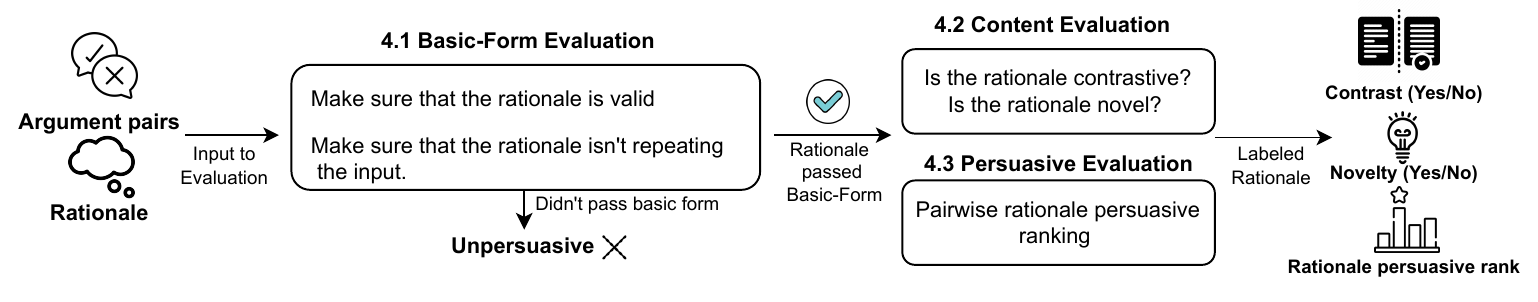}  \caption{\centering 
For the input argument pair and rationale, we filter out invalid or repetitive rationales (Section \ref{subsec:basic_form}). The qualified rationales are then analyzed based on their content (Section \ref{subsection:content_eval}) and persuasiveness (Section \ref{subsec:persuasive_eval}).
\label{fig:annotation_pipeline}}
\end{center}
\end{figure*}

\subsection{Models}
\paragraph{Considered LLMs}
Our study 
employs a set of LLMs to investigate the influence of various 
features on the generated rationales.
$(1)$ 
{\bf Open-source} models include \textit{Llama2} \cite{touvron2023llama} and \textit{Vicuna} \cite{zheng-2023-wku}, while  {\bf closed-source} models include \textit{GPT models}
(GPT-3.5-turbo and GPT4) recognized for superior performance on many downstream NLP tasks \cite{wang2023far}. 
$(2)$ \textbf{Instruction tuning} is represented by the \textit{chat} versions of Llama2 and Vicuna, where the latter is primarily fine-tuned based on human preferences between pairwise model generations. $(3)$ 
For each LLM family, we test various {\bf model sizes}, 
namely $7$-B and $13$-B versions of  \textit{Vicuna} and \textit{Llama2}  (both chat and non-chat versions) and 
\textit{Llama2-70B-chat}.\footnote{GPT4 is reported to have massive parameters, though the exact details 
are undisclosed.}

For all open-source models, we utilized 
the Hugging Face library implementations \cite{wolf2019huggingface}. As for the OpenAI models, we employed the OpenAI API \footnote{\url{https://openai.com/blog/openai-api}} to prompt both GPT-3.5-turbo and GPT4. To 
reduce randomness in generation, we set the temperature during decoding to $0$. 

\paragraph{Prompting LLMs for Ranking Arguments and Generating Rationales}
\label{subsec:pairwise_rank}
Our prompt is structured to contain three components. 
\textbf{$(1)$ System Message:} This includes a designated system setting assumed by the model during the task. 
\textbf{$(2)$ Task Description:} We describe the ranking task, assigning numerical identifiers to arguments as recommended for LLM-based ranking tasks \cite{sun2023chatgpt, pradeep2023rankvicuna}. To improve clarity, we include ranking criteria from prompts used by human annotators for assessing argument quality \cite{toledo2019automatic, gretz2020large}. Additionally, we instruct the model to generate reasoning to support its chosen argument.
\textbf{$(3)$ Formatting Examples:} We present the model with input format and the expected output. To prevent bias, we provide two formatting examples, one where argument $1$ is the winner (the selected argument in pairwise ranking) and another where argument $2$ is the winner. This ensures the model includes all expected components in its output.
\footnote{The exact prompt is included in Appendix \ref{app:ranking_prompt}.}

\begin{table*}[]
\small
\centering
\begin{tabular}{p{0.12 \textwidth}|p{0.8\textwidth}}
\hline
\textbf{Analysis Type}          & \textbf{Examples}                                                                                                                                                                                                                                                                                                                                                                                                                                                                                           \\ \hline
\textbf{Input Argument Example}     & \makecell[l]{\textcolor{blue}{\textbf{Topic: "Flu vaccination should not be mandatory"}}\\ \textbf{Argument 1:} "the state ought not intervene in individuals' right of  bodily autonomy regardless \\of societal harms. forcing vaccinations \colorbox{gray!20}{is a crude invasion to someone's body and should therefore not} \\ \colorbox{gray!20}{be done."} \\  \textbf{Argument 2:} "forcing people to participate in a system they do not trust, often with passion and for \\irrational reasons, might lead them to distrust conventional medicine as a whole, increasing the risk to \\ their health."\\ \textcolor{green}{Decision:} "Argument 1".}
                                                             \\ \hline
\textbf{Invalid reasoning } &  \textcolor{red}{\textbf{Reasoning:} "argument 1 is more convincing than argument 2}\\ \hline
\textbf{Repetitive reasoning} &  \textcolor{red}{\textbf{Reasoning:} "It  \colorbox{gray!20}{ is a crude invasion to someone's body and should therefore not be done."}}\\ \hline


\textbf{Valid reasoning} & \textcolor{purple}{\textbf{Reasoning:} "The first argument emphasizes the importance of bodily autonomy and the limitation of state intervention in personal choices.  \colorbox{green!20}{It presents a clear ethical principle that can be applied consistently across different situations.} 
\colorbox{blue!20}{The second argument highlights potential negative consequences of mandatory vaccination.}, \colorbox{blue!20}{which, while valid, may not be as persuasive as the ethical principle of bodily autonomy.}}"\\ \hline

\end{tabular}
\caption{\centering Examples of \textit{Invalid Reasoning} (second row), \textit{Repetitive Reasoning} (third row) from non-compliant sets, and \textit{Valid Reasoning} (fourth row) with highlighted text: \colorbox{gray!20}{Repetitive part}, \colorbox{blue!20}{Contrastive part}, \colorbox{green!20}{Novel part}. \label{tab:analysis_examples}}
\end{table*}

\section{Rationale Evaluation}

Figure \ref{fig:annotation_pipeline} outlines our evaluation process, which consists of three key stages:  $(1)$ \textbf{Basic-Form Evaluation:} This initial stage filters out meaningless rationales, ensuring only valid ones proceed for further analysis, similar to the concept of surface-form evaluation \cite{joshi2023machine}. $(2)$ \textbf{Content Evaluation:} We assess the rationale's content by analyzing its support through \textit{contrast} and its informativeness through \textit{novelty}, aiming to understand how rationale content influences its persuasiveness.  $(3)$ \textbf{Persuasiveness Evaluation:} We assess the rationale's persuasiveness relative to other generated rationales supporting the chosen argument.

We rely on human annotators to evaluate each stage, using the $270$ rationale subset from the IBM-9k described in Table \ref{tab:dataset_summary}. This annotated set is used to: (1) Analyze the influence of rationale content on the rationale persuasiveness, and (2) Explore automatic persuasiveness evaluation methods to reduce the cost of human evaluation, especially in utility-driven tasks \cite{joshi2023machine}.

We use Mechanical Turk workers for annotations at each evaluation stage. Each dimension in basic form and content evaluation is assessed with yes/no questions, with the majority vote of three workers determining the final decision on contrast and novelty. For persuasiveness, rationales are evaluated through pairwise comparisons, where three workers vote on whether a rationale is more, less, or equally persuasive compared to another. Workers also write $1-2$ sentences explaining their choices to ensure critical thinking. Rigorous quality assurance methods ensure annotation reliability.
\footnote{Appendix \ref{app:hits}
 details our annotation guidelines, quality assurance methods, and annotator reliability.}



\subsection{Basic-Form Evaluation}
\label{subsec:basic_form}
We examine two aspects of the quality of a rationale to assess the rationale form:
\textit{Validity}: Is the rationale grammatically correct and coherent?
\textit{Repetition}: Does the rationale merely reiterate the input argument, either fully or in summary, without adding any new insight or justification?

Examples of rationales that fail to meet these criteria are provided in Table \ref{tab:analysis_examples}, with invalid rationales shown in the second row and repetitive rationales in the third row. If a rationale doesn't meet these basic-form requirements, it is disregarded from further evaluation and deemed unpersuasive by default.



\subsection{Content Evaluation}
\label{subsection:content_eval}

For \textit{contrast}, we assess the LLM's ability to refute the argument it did not choose. Our goal is to determine if refuting the alternative argument enhances the rationale's persuasiveness. For \textit{novelty}, we evaluate whether the rationale introduces new information or a new perspective not explicitly mentioned in the arguments, thereby increasing its persuasiveness. An example of a valid rationale with highlighted contrastive and novel (new perspective) parts can be found in Table \ref{tab:analysis_examples}, row 4\footnote{We also analyzed rationale content for support by evaluating \textit{association} \cite{wiegreffe2021measuring, wiegreffe2021teach}, determining if the rationale highlights key points in the chosen argument. Most LLMs supported their choices through association, offering no unique information for persuasiveness ranking.}.

\subsection{Persuasiveness Evaluation}
\label{subsec:persuasive_eval}
In recommendation tasks, persuasive explanations help users understand why a certain item or choice is recommended, convincing them to accept it \cite{wang2014also,tran2023user}. Similarly, in argument ranking, persuasiveness of the rationales can be defined as the ability to convincingly justify the model's recommendation of one argument over another. Due to the subjective nature of this task, we opted against assigning a single persuasiveness score. Instead, we evaluate persuasiveness through pairwise comparisons, allowing us to assess the persuasiveness abilities of different models supporting the same choice.

\paragraph{Human Evaluation of Persuasiveness}
Due to the quadratic nature of pairwise comparisons, we randomly select one third of the rationale pairs for persuasion described in Table \ref{tab:dataset_summary}, resulting in $360$ pairs. After excluding rationales that do not meet basic quality standards, we are left with $204$ pairs for human annotations. We refer to this subset as IBM-9k (annotate set).



\begin{table*}[htbp]
\centering
\small
\resizebox{\textwidth}{!}{
\begin{tabular}{@{}lccccccc@{}}
\toprule
\multirow{4}{*}{\textbf{Model}} & \multicolumn{2}{c}{\multirow{3}{*}{\textbf{IBM9k (Annotated Set)}}} & \multirow{3}{*}{\textbf{IBM9k (Full Pairs)}} & \multicolumn{4}{c}{\textbf{IBM-30k-rank}} \\ 
\cmidrule(lr){5-8}
& \multicolumn{2}{c}{} & & \multicolumn{4}{c}{\textbf{Quality Differences}} \\ 
\cmidrule(lr){5-8}
& \multicolumn{2}{c}{} & & \textbf{Full Pairs} & \textbf{0-0.25} & \textbf{0.25-0.5} & \textbf{0.5-1.} \\ 
\cmidrule(lr){2-8}
& \textbf{APR ($\delta$) with Human-Eval $\uparrow$} & \textbf{APR ($\delta$) GPT4 Eval $\uparrow$} & \textbf{APR ($\delta$) GPT4 Eval $\uparrow$} & \multicolumn{4}{c}{\textbf{APR ($\delta$) GPT4 Eval $\uparrow$}} \\ 
\midrule
\textbf{Llama2-13B-Chat} & $2.28 (0.48)$ & $2.14 (0.37)$ & $3.42 (1.95)$ & $3.75 (1.74)$ & $3.69 (1.73)$ & $3.85 (1.69)$ & $3.60 (2.02)$ \\ 
\textbf{Llama2-7B-Chat} & $3.14 (1.46)$ & $3.42 (0.78)$ & $3.85 (2.20)$ & $4.15 (2.09)$ & $4.11 (2.07)$ & $4.43 (2.13)$ & $3.66 (2.12)$ \\ 
\textbf{Vicuna-7B} & $3.63 (0.80)$ & $4.18 (1.72)$ & $4.39 (1.49)$ & $3.75 (1.36)$ & $3.61 (1.38)$ & $3.75 (1.28)$ & $4.60 (1.24)$  \\ 
\textbf{Vicuna-13B} & $4.36 (1.56)$ & $3.72 (1.10)$ & $4.67 (1.46)$ & $4.45 (1.41)$ & $4.60 (1.37)$ & $4.17 (1.41)$ & $4.45 (1.55)$  \\ 
\textbf{GPT-3.5-Turbo} & $5.18 (1.16)$ & $6.00 (1.48)$ & $6.14 (1.53)$ & $5.08 (1.37)$ & $5.11 (1.35)$ & $4.95 (1.53)$ & $5.00 (1.00)$  \\ 
\textbf{GPT4} & $5.72 (1.55)$ & $5.72 (1.19)$ & $5.92 (1.27)$ & $5.82 (1.06)$ & $5.86 (0.93)$ & $5.82 (1.18)$ & $5.66 (1.34)$ \\ 
\textbf{Llama2-70B-Chat} & \textbf{7.00} \textbf{(1.09)} & \textbf{6.18} \textbf{(1.77)} & \textbf{6.57} \textbf{(1.66)} & \textbf{6.29} \textbf{(0.98)} & \textbf{6.14} \textbf{(1.07)} &  \textbf{6.09} \textbf{(1.71)} & \textbf{5.91} \textbf{(0.91)}  \\ 
\bottomrule
\end{tabular}
}
\caption{\centering Average Persuasive Rank (APR) ($\delta$) for $7$ instruction-tuned LLMs and datasets. $\delta$ denotes the standard deviation. $\uparrow$ indicates higher persuasiveness. Rows are sorted by Human-Eval APR in ascending order.}
\label{tab:results}
\end{table*}

\paragraph{Automatic Evaluation of Persuasiveness}
To assess persuasiveness rankings on a larger scale across all pairs in our study (both the full IBM-9k and IBM-30k pairs), we utilize GPT4 for automatic persuasiveness ranking. GPT4 is selected for its proven effectiveness in evaluating various downstream tasks \cite{liu2023g, chiang-lee-2023-large}. We benchmark GPT4's rankings against human persuasiveness rankings on the annotated set and then report its persuasiveness ranking scores across all IBM-9k rationale pairs (IBM-9k Full Pairs) and the IBM-30k dataset.

\paragraph{Persuasive Ranking Metric}
For both human and automatic evaluations, we use the scoring formula proposed by \citet{qin2023large} in ranking passages for retrieval tasks, to rank persuasiveness of the rationales.
The score  $s_i$  for a rationale $r_i$ is given by:

\begin{equation}
    \label{eq:first}
    s_{i} = 1 \cdot \sum_{\substack{j=1 \\ j \neq i}}^{M} \mathbb{I}_{r_i > r_j} + 0.5 \cdot \sum_{\substack{j=1 \\ j \neq i}}^{M} \mathbb{I}_{r_i = r_j}
\end{equation}
where $M$ is the total number of considered models and $r_i$ and $r_j$ are the rationales from model $i$ and model $j$, respectively. This formula adds $1$ to the score $s_i$ if a rationale $r_i$ is considered more persuasive than $r_j$, and $0.5$ if it is considered equally persuasive. To determine the overall persuasiveness of each model, we use the $s_i$ scores to rank the models' generated rationales for each argument pair and report the \textbf{Average Persuasiveness Rank (APR)} of each model as the final persuasiveness score, ranging from $1$ ranked the least persuasive and $M$, which is the total number of models included in the comparison, as the most persuasive.

To compute Equation \ref{eq:first} using GPT4, we instruct the model to compare the persuasiveness of rationale $1$ and rationale $2$ in supporting the argument. Same as human evaluation, we include a third option for GPT4 to select if it finds both rationales equally persuasive. Furthermore, following the method described by \citet{qin2023large}, we present the rationale pairs to GPT4 twice, each time with the order of rationales switched. If GPT4's decision differs between the two prompts, we consider the rationales to be equally persuasive and increase the $s$ score of each rationale by $0.5$ \footnote{Prompt in Appendix \ref{app:persuasion_eval_gpt4}.}.

\section{Results and Analysis}

\subsection{Persuasiveness Rankings of Rationales}


\paragraph{Human and Automatic Persuasive Rankings (RQ1, RQ2)}

Table \ref{tab:results} presents the APR in all data sets. Llama2-7B and Llama2-13B were excluded from the rankings because their basic-form annotations indicated a consistent failure in quality check, making them the least persuasive by default\footnote{Appendix \ref{app:characteristics} details the basic-form distribution across all models.}. Therefore the APR is reported across $7$ LLMs instead of $9$.
Llama2-70B-chat consistently generated the most persuasive rationales. This was evident in both human and automatic rankings with GPT4, surpassing even closed-source GPT models. This result highlights the potential of open-source models like Llama2-70B-chat in tasks such as pairwise argument ranking.


For the IBM-9k annotated set, GPT4 did not perfectly match the APR with human evaluation. However, GPT4 agreed with human evaluation in the persuasiveness ranking order of the included LLMs, except for the rankings of GPT4 vs. GPT-3.5-Turbo and Vicuna-7B vs. Vicuna-13B. This suggests that GPT4 can differentiate between the persuasiveness of rationales when differences are significant, but may disagree with human judgment when the persuasiveness scores are close.

\begin{figure}[ht]
    \includegraphics[width=0.5\textwidth]{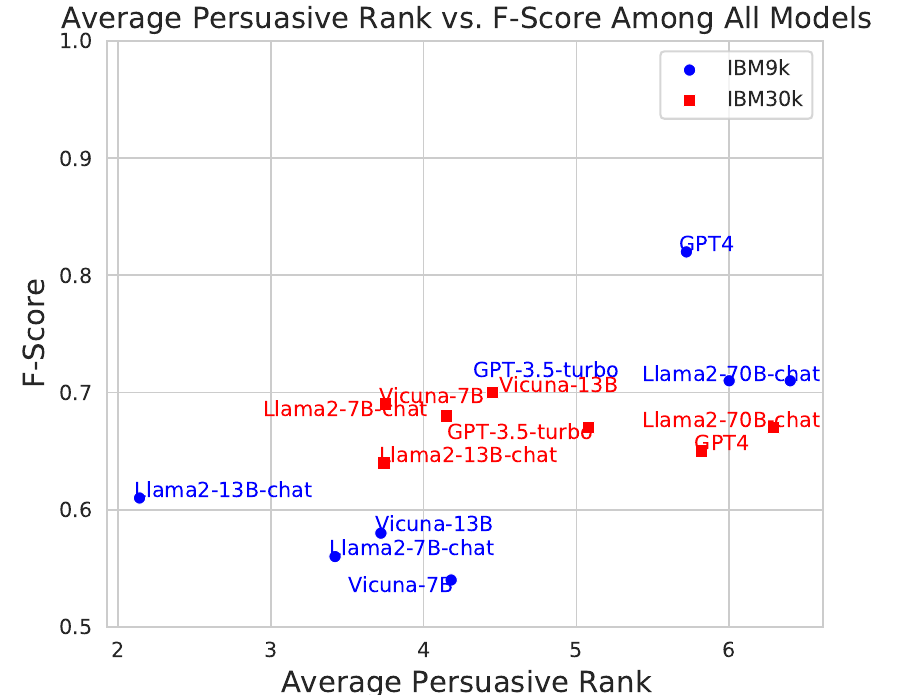}
    \centering
    \caption{\centering 
  Persuasion Ranking vs F-score
    \label{fig:rm_fscore}}
\end{figure}

For the IBM-30k data set, the variation in the difference in the quality of arguments had a limited effect on the persuasiveness of the rationale. The rationale generated by Llama2-70B chat remained the most persuasive, followed by those of GPT4 and GPT-3.5-turbo. This indicates that different LLMs tend to follow a similar rationalization strategy regardless of the quality difference. For all datasets, we found that instruction tuning and model size improves persuasiveness
\footnote{Details are in Appendix \ref{app:model_persuasiveness_char}.}.



\subsection{What contributes to the rationale persuasiveness? (RQ3)}

\paragraph{Model Accuracy $\neq$ Rationale Persuasion}
Figure \ref{fig:rm_fscore} shows that the LLM's ability to accurately predict the annotated higher-ranked argument, measured by the F1 score between the LLM's predicted argument and the annotated argument on the full unfiltered argument pairs of the IBM-9k and IBM-30k datasets, does not necessarily correlate with higher persuasiveness scores measured by GPT4 across the IBM-9k annotated set and the IBM-30k full set. This is further supported by the insignificant Pearson correlation results, with $p > 0.05$ for both datasets.

For example, despite having the highest persuasiveness rank, Llama2-70B-chat falls behind GPT4 in the F1 score for the IBM-9k dataset. This trend is more apparent with the IBM-30k pairs, where both GPT4 and GPT-3.5-turbo have lower F1 scores compared to the Vicuna models, yet achieve higher persuasive rankings. The drop in F1 scores can be attributed to the quality variation in the IBM-30k test set, affecting the LLM's ability to agree with the annotated higher argument, but having limited impact on how the model supports its prediction. These observations indicate that a model's ability to convincingly support an argument extends beyond mere accuracy in predicting the labeled argument, suggesting a complex interplay of factors that influence a model's persuasive capabilities.

\begin{figure}[ht]
    \centering
    \includegraphics[width=0.5\textwidth]{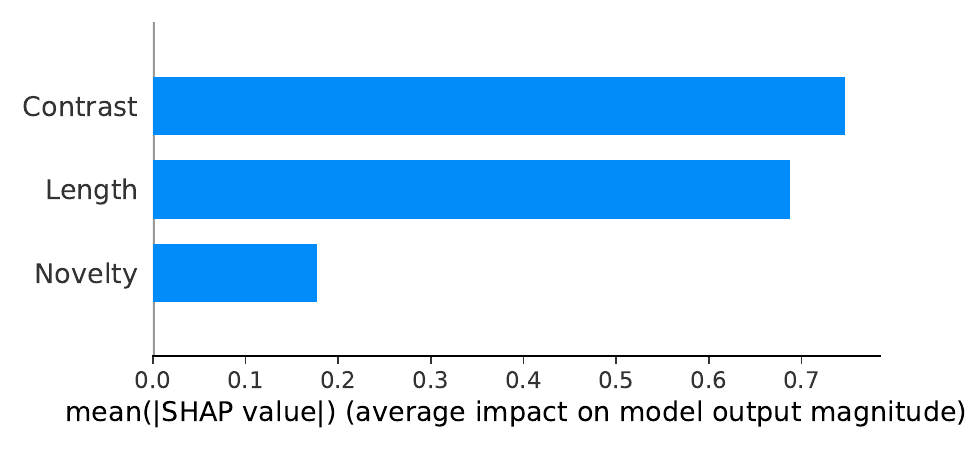}
    \centering
    \caption{\centering 
   \textit{SHAPLey} values of each feature. The higher the value , the higher the impact on average persuasiveness rank.
    \label{fig:persuasion_shap}}
\end{figure}

\paragraph{Rationale Content Analysis}
In addition to \textit{Contrast} and \textit{Novelty}, we also explore the observable characteristic of rationale \textit{Word length} on the  persuasiveness ranking of the rationales and investigate the role of these attributes. 
We formulate this as a regression task, employing a \textit{random forest regressor} ($f$) to predict persuasiveness ranking based on the features: length ($X_{length}$), contrast ($X_{contrast}$), and novelty ($X_{novelty}$).
$Ranking = f(X_{length}, X_{contrast}, X_{novelty})$. We convert the contrast and novelty majority votes for each rationale into binary values. 
Upon estimating $f$, we use the \textbf{SHAP explainer} \cite{lundberg2017unified} to determine the impact of each feature on the persuasiveness ranking. We particularly used SHAP as it takes into consideration the feature interaction when estimating the individual feature impact on the predictions.

 Figure \ref{fig:persuasion_shap} shows that \textit{contrast} is the most influential factor in persuasiveness. This aligns with studies advocating for contrastive explanations in truth verification \cite{si2023large} but deviates from \citet{joshi-etal-2023-machine}, where contrast had minimal influence on rationale utility. We hypothesize that this is intuitive, given the nature of our task.  By weakening the alternative arguments, we can make the argument choice more acceptable and enhance the rationale's persuasiveness. \textit{Length} is also significant, indicating that more detailed explanations may improve persuasiveness. Lastly, \textit{novelty} has a less pronounced impact, suggesting that while new information is valuable, its role is secondary to contrast and length in this context. 

\begin{figure}[ht]
    \centering
    \small
    \includegraphics[width=.45\textwidth]{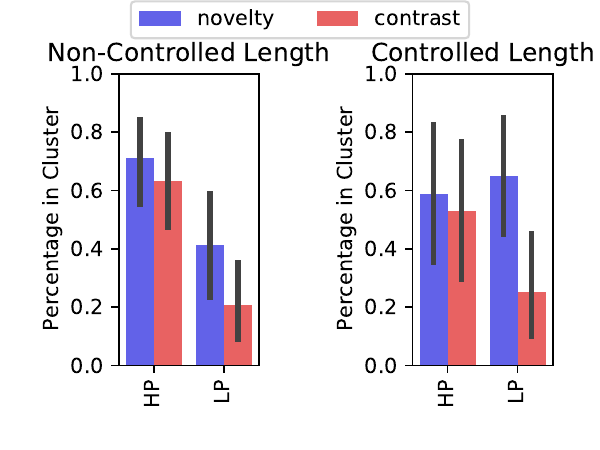}
    \centering
    \caption{\centering 
   \textit{Contrast} and \textit{Novelty} \% in different categories of rationale rating.
    \label{fig:control_vs_noncontrol}}
\end{figure}

To understand the content contribution independent of content length, we cluster rationales into two groups: \textit{High-Persuasive (HP)} and \textit{Low-Persuasive (LP)} clusters, using k-means clustering. We then control for length variations by focusing only on rationales with word lengths within $20\%$ of each other. This ensures that any observed differences in persuasiveness are primarily due to content, not length.
Figure \ref{fig:control_vs_noncontrol} illustrates that in the IBM-9k annotated set of rationales
, both novelty and contrast percentages are significantly higher (ANOVA-test, $p<0.05$) in the High-Persuasive group. However, in the controlled length rationale set, only contrast exhibits a significant increase in the High-Persuasive group (ANOVA-test, $p<0.05$). These results verify the SHAP analysis, emphasizing the importance of contrast in persuasion. Conversely, the presence of novelty in lengthy rationales may act as a confounding factor, potentially inflating its significance.

\subsection{Controlling Persuasiveness (RQ4)}

We aim to use the insights from the previous session to improve the model's ability to generate persuasive rationales. We experimented with \textbf{Re-prompting the LLM:} This involved asking the model to provide two sentences supporting its chosen argument and two sentences refuting the alternative argument. The goal was to encourage the model to include contrastive rationales with sufficient length,  proven influential for persuasiveness. we compare this method against 
\begin{table}[htbp]
\centering
\small
\begin{tabular}{@{}lccc@{}}
\toprule
\textbf{Model} & \textbf{APR GPT4 Eval $\uparrow$} \\ 
\midrule
\textbf{Llama2-7B-Chat} & $4.31(2.86)$ \\ \midrule
\textbf{\textit{Llama2-7B-chat-persuasion-prompted}} & $6.65(2.97)$ \\
\textbf{\textit{Llama2-7B-chat-persuasion-refined}} & $5.15(2.88)$ \\
\midrule
\textbf{Llama2-13B-Chat} & $3.68(2.00)$ \\ 
\textbf{Llama2-70B-Chat} & \textbf{7.89}  \textbf{(2.05)} \\ 
\textbf{Vicuna-7B} & $5.57(1.74)$ \\ 
\textbf{Vicuna-13B} & $5.52(2.06)$ \\ 
\textbf{GPT-3.5-Turbo} & $7.63(1.53)$ \\ 
\textbf{GPT4} & $7.21(1.39)$ \\ 
\bottomrule
\end{tabular}
\caption{\centering (APR)  LLMs on the IBM9k (Full Pairs) dataset using GPT4. \textit{Italicized} rows indicate the Llama2-7B-chat models experimented for enhanced persuasiveness.}
\label{tab:results_ibm9k_pers}
\end{table}
\textbf{Evaluate and Refine:} which is a form of self-refinement \cite{huang2022large}. The model first assesses whether the generated rationale was persuasive. If the model determines that the rationale is not persuasive, it then generates a more persuasive one. Both methods were applied to the Llama2-7B-chat model, which, as shown in Appendix \ref{app:characteristics}, had a low rate of generating contrastive rationales. We refer to the new rationales generated by the model as \textit{Llama2-7B-chat-persuasion-prompted} and \textit{Llama2-7B-chat-persuasion-refined}, respectively \footnote{Prompts are in Appendix \ref{app:persuasion_improved}.}.




Table \ref{tab:results_ibm9k_pers} shows that \textit{Llama2-7B-chat-persuasion-prompted} ranks higher in persuasiveness with GPT4-based ranking compared to both Llama2-7B-chat and self-refined rationales (\textit{Llama2-7B-chat-persuasion-refined}), which emphasizes the importance of contrast and detail in enhancing rationale persuasiveness. However, the new rationales still lag behind Llama2-70B-chat and GPT models, indicating that larger models may rely on persuasive factors unexplored in our work. \textit{Evaluate and Refine} method did not improve persuasiveness compared to prompting with persuasive parameters, suggesting that LLMs benefit more from alignment on persuasive factors.

\section{Conclusion and Future Work}
This paper presents a comprehensive analysis of the persuasiveness of free-text rationales generated by various LLMs. Our results show that open-source models, particularly Llama2-70B-chat, generate highly persuasive rationales, surpassing strong closed-source GPT models. While GPT4's rankings generally align with human judgments, discrepancies arise due to the task's inherent subjectivity.
We proposed a detailed human evaluation studying key factors contributing to persuasiveness. We found that  \textit{contrastive rationales},where the model justifies its choice and refutes the alternative, the most significant. We also demonstrated that prompting models with specific persuasiveness parameters enhances rationale persuasiveness. Future work will explore the user acceptance of model-chosen arguments and investigate other subjective tasks beyond pairwise argument ranking.

\section{Limitations}
This study primarily utilized rationale evaluation taxonomies to assess persuasiveness. Future work could incorporate additional factors from persuasive theory to gain a deeper understanding of what different LLMs rely on to support their choices. Our annotated sample size is relatively small, as we prioritized quality control over a larger quantity of annotations. Although we hypothesize that our results would be consistent with a larger sample, it would strengthen our findings to re-evaluate our methods on a broader dataset. Additionally, expanding the study to other domains where the task is inherently subjective, beyond pairwise argument ranking, would provide a more comprehensive evaluation.

\section{Ethical Statement}
Persuasive rationales can enhance transparency, particularly in subjective tasks, by making recommendations more acceptable to users. However, there is a potential ethical concern that persuasive rationales could be used adversarially  to promote biased or nonfactual arguments. Therefore, it is crucial to consider the ethical implications of deploying persuasive rationales and to develop safeguards to prevent misuse.

\bibliography{anthology,custom}

\appendix

\section{Argument Pairs Agreement Distribution }
\label{app:agreed_upon}

\begin{figure}[ht]
    \includegraphics[width=0.45\textwidth]{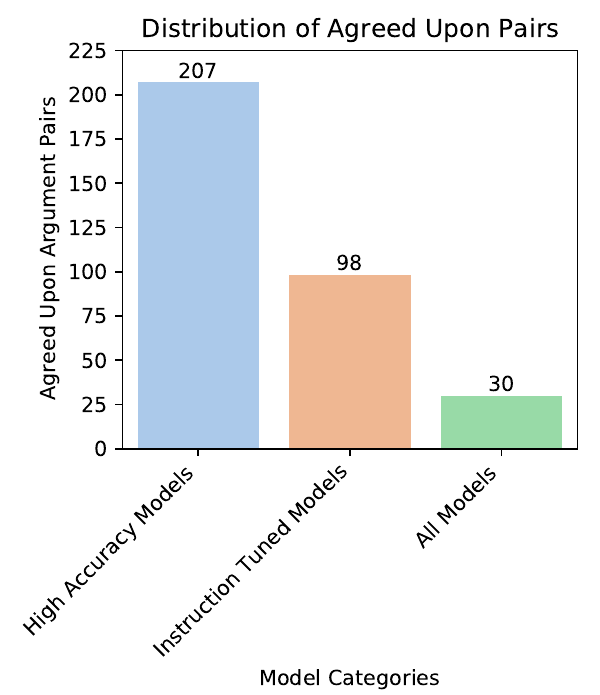}
    \centering
    \caption{\centering 
  Distribution of argument pairs across different categories of models in the IBM-9k sampled set. 
    \label{fig:arg_dist}}
\end{figure}

Figure \ref{fig:arg_dist} illustrates that the number of agreed-upon argument pairs decreases as more models are included in the analysis. The "High Accuracy" category includes GPT-4, GPT-3.5-turbo, and Llama2-70B-chat. The "Instruction Tuned" category adds the remaining instruction-tuned models to the high-accuracy models: Llama2-7B-chat, Llama2-13B-chat, Vicuna-13B, and Vicuna-7B. Finally, the "All Models" category includes the non-instruction-tuned models Llama2-7B and Llama2-13B in addition to those in the previous categories. For a more comprehensive analysis, we included all models in our analysis.

\paragraph{Obtaining Rationalization Pairs}
For each argument pair, we generate $9$ different rationales from the included LLMs. Using pairwise comparisons to rank these rationales results in $36$ combinations per argument pair. Consequently, for the total filtered argument pairs, we have $1080$ rationale pairs for the IBM-9k dataset and $5184$ for the IBM-30k dataset.

\section{Pairwise Ranking prompt}
\label{app:ranking_prompt}

Table \ref{tab:first_stage} shows the exact prompt used for our first stage pairwise ranking.  
The "Expected Output" section of the prompt indicates the format in which the model generates responses and not an actual output.
\lstdefinelanguage{json}{
  basicstyle=\small\ttfamily,
  numbers=left,
  numberstyle=\tiny,
  stepnumber=1,
  numbersep=8pt,
  showstringspaces=false,
  breaklines=true,
  frame=lines,
  backgroundcolor=\color{lightgray},
  literate=
    *{0}{{{\color{blue}0}}}{1}
    {1}{{{\color{blue}1}}}{1}
    {2}{{{\color{blue}2}}}{1}
    {3}{{{\color{blue}3}}}{1}
    {4}{{{\color{blue}4}}}{1}
    {5}{{{\color{blue}5}}}{1}
    {6}{{{\color{blue}6}}}{1}
    {7}{{{\color{blue}7}}}{1}
    {8}{{{\color{blue}8}}}{1}
    {9}{{{\color{blue}9}}}{1}
    {:}{{{\color{red}{:}}}}{1}
    {,}{{{\color{red}{,}}}}{1}
    {\{}{{{\color{red}{\{}}}}{1}
    {\}}{{{\color{red}{\}}}}}{1}
    {[}{{{\color{red}{[}}}}{1}
    {]}{{{\color{red}{]}}}}{1},
}

\begin{table*}
\small
\begin{center}
  \begin{tabular}{|>{\centering\arraybackslash}p{0.9\textwidth}|}
    \hline
    \textbf{Pairwise Ranking Prompt} \\
    \hline
    \textbf{System message}
    You possess the art of argumentation. \\
    \textbf{Task definition}
    You will receive two arguments, each identified by a numerical identifier [] and a Topic.  \textit{Disregarding your own opinion on the topic, given the arguments, the human decision, and the human reasoning, decide which argument you would recommend.} Choose argument [1] if you recommend argument [1] over argument [2]. Choose argument [2] if you recommend argument [2] over argument [1]. Format your output in a JSON format with "decision" and "reasoning" keys. \\
    \hline
    \textbf{Reminder:} Make sure to choose only one argument and provide a convincing reasoning why you choose this argument over the other one. Generate only the JSON output with decision and reasoning, do not generate any additional thought process or discussion. \\
    \hline
    \textbf{Formatting Examples}\\
    Example 1:
    \begin{lstlisting}[language=json]
{
  "topic": "topic 1",
  "1": "argument 1",
  "2": "argument 2",
  "model_decision": "argument 1",
  "model_reasoning": "reason model chose argument 1"
}
    \end{lstlisting}

    Output: 
    \begin{lstlisting}[language=json]
{
  "decision": 1,
  "reasoning": "reason for choosing argument 1"
}
    \end{lstlisting}

    Example 2 \\
    \begin{lstlisting}[language=json]
{
  "topic": "topic 2",
  "1": "argument 1",
  "2": "argument 2",
  "model_decision": "argument 1",
  "model_reasoning": "reason model chose argument 1"
}
    \end{lstlisting}

Output \\
    \begin{lstlisting}[language=json]
{
  "decision": 2,
  "reasoning": "reason for choosing argument 2"
}
    \end{lstlisting}

    \vspace{0.5cm}

    \textbf{Annotation Example} \\
    \begin{lstlisting}[language=json]
{
  "topic": "{}",
  "1": "{}",
  "2": "{}",
  "model_decision": "argument {}",
  "model_reasoning": "{}"
}
    \end{lstlisting}

   Expected Output (generated by the model in json format) \\
    \begin{lstlisting}[language=json]
{
  "decision": "...",
  "reasoning": "..."
}
    \end{lstlisting} \\
    \hline
  \end{tabular}
\end{center}
\caption{\centering Pairwise argument ranking prompt. \textit{italicized} part in \textbf{Task definition} is the prompt given to human annotators described in \cite{gretz2020large,toledo2019automatic} .  \label{tab:first_stage}}
\end{table*}

\section{Mechanical Turk HITS}
\label{app:hits}
\subsection{Basic-form Evaluation in Detail}

\paragraph{Evaluation Process and Worker Reliability}

 We employ Mechanical Turk workers with more than $95\%$ approval rate and more than $5000$ approved HITS. 
 Workers are instructed to select \textbf{yes} if the rationale is both valid and devoid of any partial or full repetition of the chosen argument. If the rationale violates either of these conditions, Turkers are directed to choose \textbf{no}. Additionally, they are required to specify the reason for rejecting the rationale, selecting between "invalid rationale" or "repetitive rationale."
At first, workers were given $20$ examples to help them understand the task requirements and estimate its difficulty. Along with the task description, clear instructions and examples were provided to avoid any possible confusion. Three workers evaluated each sample. The reliability of the workers was measured using Krippendorff's alpha ($\alpha$) \cite{krippendorff2011computing}. The initial score of $0.53$ was achieved for basic labeling criteria and $0.27$ for identifying reasons for non-compliance. To improve the evaluation quality, we disqualified workers who failed to answer hidden test questions and introduced a set of $20$ examples with revised guidelines. This led to an improved score of $0.80$ for basic form labeling and $0.66$ for identifying reasons for non-compliance on the additional set of $20$ examples. Using these revised guidelines, we evaluated the final set of $270$ rationales. The reliability score for this phase was $0.76$ for basic form labeling and $0.71$ for identifying reasons for failure, whether due to validity or repetition (non-compliance). The majority votes from workers' assessments were used to evaluate each sample. Samples that failed to meet basic form criteria, as determined by the majority vote, were excluded from further evaluation phases.

\begin{figure*}[ht]
    \centering
    \includegraphics[width=\textwidth]{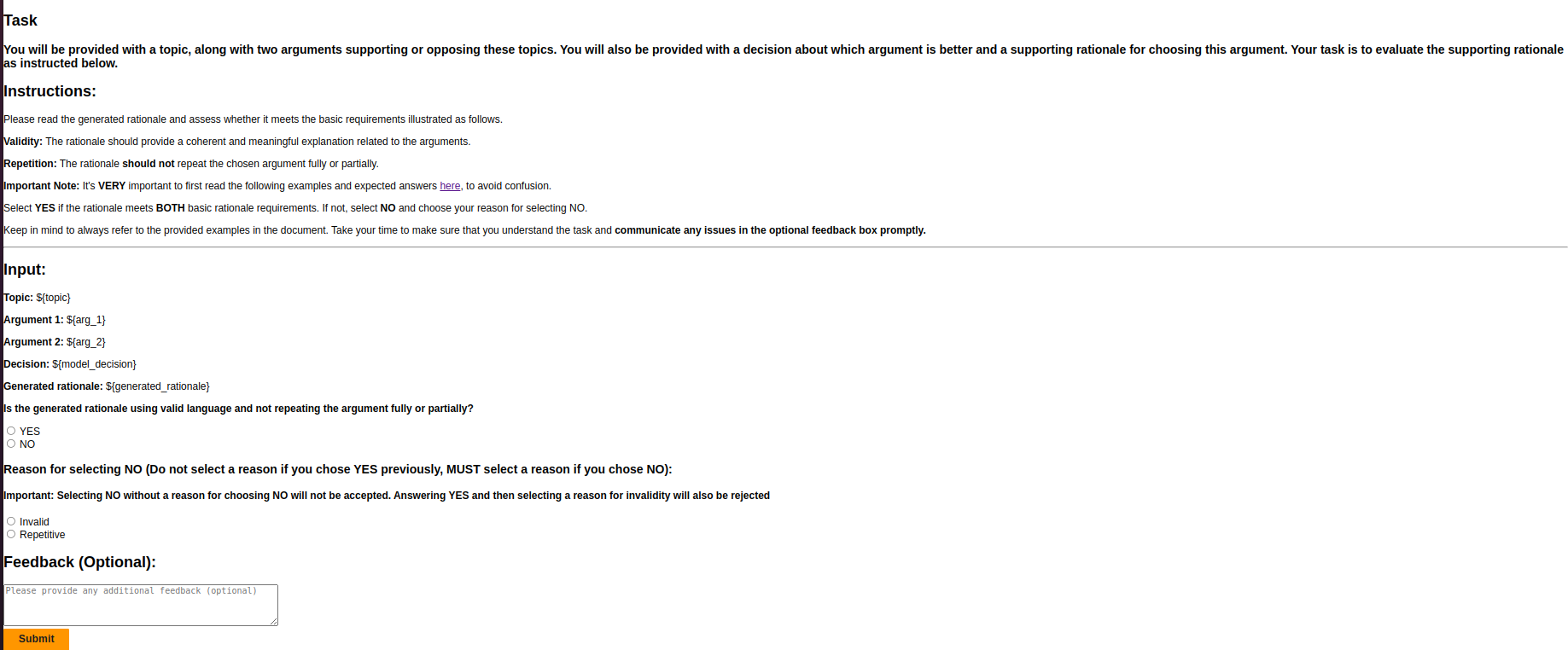}
    \centering
    \caption{\centering 
  A screenshot form basic-form MTurk HIT for basic-form evaluation.
    \label{fig:basic-form-hit}}
\end{figure*}

\paragraph{Basic-form HIT}
\label{app:basic-form-hit}
Figure \ref{fig:basic-form-hit} shows the actual MTurk HIT given to Turkers to evaluate the basic form. First, workers are asked to select YES/NO based on the validity and repetition criteria. If they select NO, they are asked to choose a reason between \textbf{Invalid} and \textbf{Repetitive} for selecting NO.

\subsection{Content Evaluation HITs}
\label{app:content_eval}

\paragraph{Annotator Qualification Process }
Similar to the basic-form evaluation, we conduct this step using YES/NO questions to determine whether the rationale is contrastive or novel. These questions are answered by proficient English-speaking Mechanical Turk workers who have passed our qualification test.
Content evaluation began with a qualification task for our annotators, all of whom are proficient in English. This initial task consisted of annotating $10$ sample rationales. The samples were selected based on their known, expected annotations in novelty and contrast to ensure the accuracy of the qualification process. Each sample was reviewed by $5$ workers. Only those workers who accurately completed at least $8$ out of the $10$ questions and achieved more than $90\%$ agreement with the expected annotations were retained for the subsequent evaluation.

\paragraph{Final Content Evaluation }

For each sample, we employ three qualified workers to assess both \textit{contrast} and \textit{novelty} aspects, using a binary YES/NO selection. The final label for each rationale is determined by the majority vote among these workers. For the complete final evaluation set, we computed Krippendorff's alpha coefficient, resulting in a value of $0.82$ for contrast, indicating a high level of annotator agreement, while it stood at $0.31$ for novelty, suggesting a relatively lower agreement. We attribute this discrepancy to the complexity of determining whether certain information constitutes a novel viewpoint or not 
\footnote{ Experiments with random workers (with over $95\%$ approval rate and over $5,000$ approved HiTs) on the same subset yielded Krippendorff's alpha values of $0.17$ and $0.18$ for contrast and novelty, respectively. These findings emphasize the importance of our qualification process in obtaining reliable annotations.}.

\begin{figure*}[ht]
    \centering
    \includegraphics[width=\textwidth]{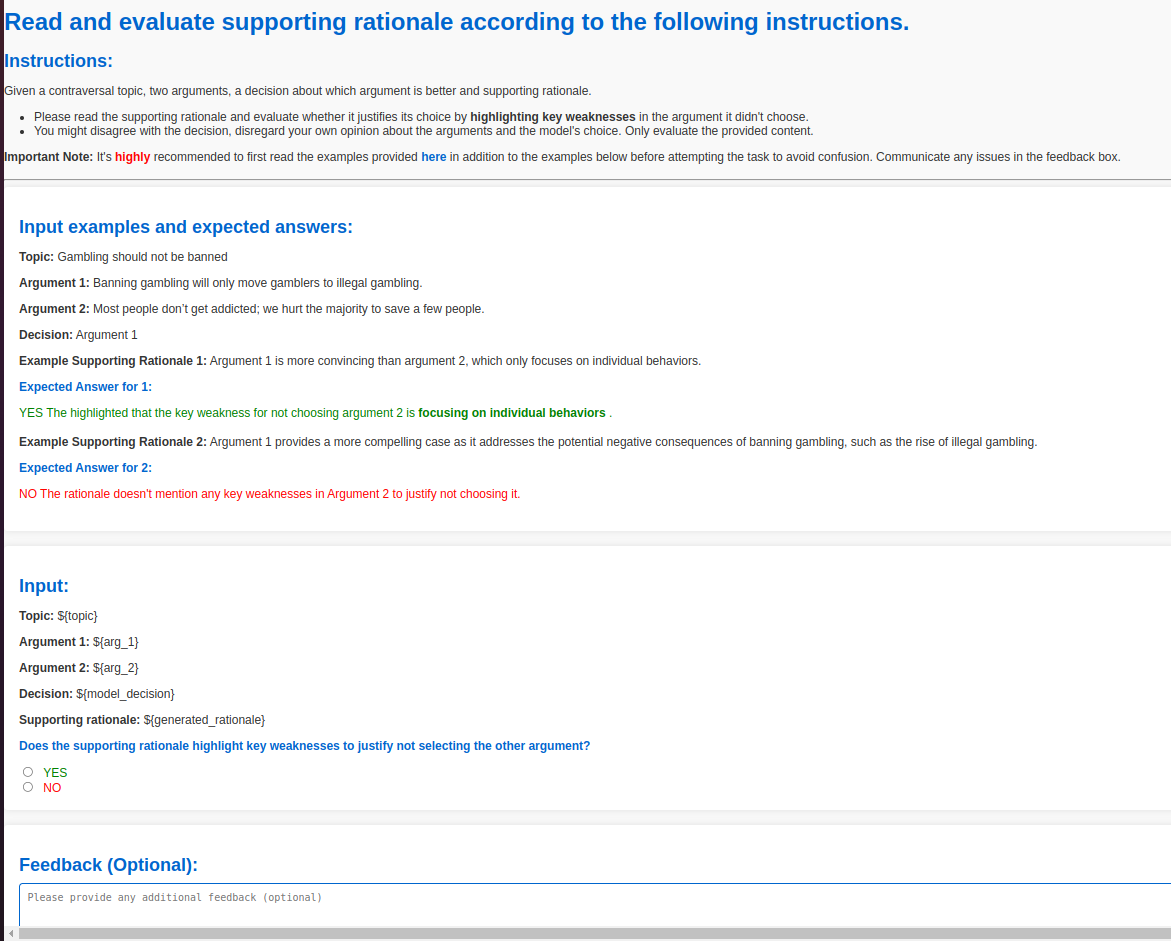}
    \centering
    \caption{\centering 
  A screenshot form  MTurk HIT for contrast evaluation.
    \label{fig:contrast-eval}}
\end{figure*}

\begin{figure*}[ht]
    \centering
    \includegraphics[width=\textwidth]{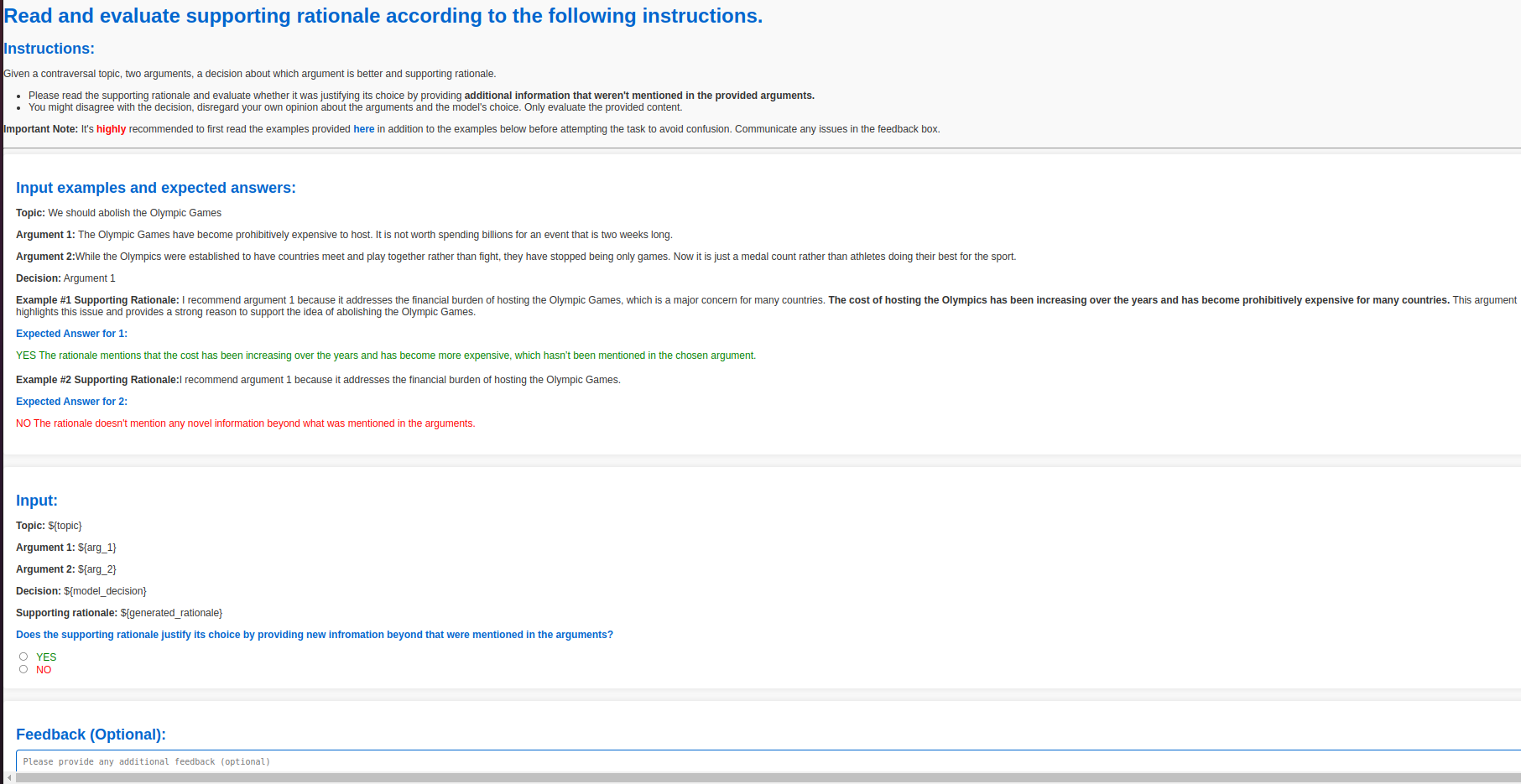}
    \centering
    \caption{\centering 
  A screenshot form  MTurk HIT for novelty evaluation.
    \label{fig:novelty-eval}}
\end{figure*}

Figure \ref{fig:contrast-eval} shows the Mechanical Turk HIT given to Mechanical Turk workers to evaluate contrast while  figure \ref{fig:novelty-eval} shows the Mechanical Turk HIT given to Mechanical Turk workers to evaluate novelty. 

\subsection{Persuasiveness Evaluation Details  and HIT Guidelines}
\label{app:persuasion_hitt}

To verify the clarity and efficacy of our instructions, we present workers with a set of $10$ pairs selected from distinct topics. Five of these pairs exhibit significant differences in rationale form, including variations in length and level of detail, while the remaining five pairs are comparable in lengths. We intentionally provide easier examples to ensure that workers follow the guidelines. Annotators had perfect agreement for the set where rationales varied significantly. For the comparable rationale pairs, the interannotator reliability, as measured by Krippendorff's Alpha, reached $0.55$.
The inter-annotator reliability for the full set, reached $0.64$. 
We use these annotation guidelines to obtain the final persuasion set, achieving a Krippendorff's Alpha score of $0.56$.

\begin{figure*}[ht]
    \centering
    \includegraphics[width=\textwidth]{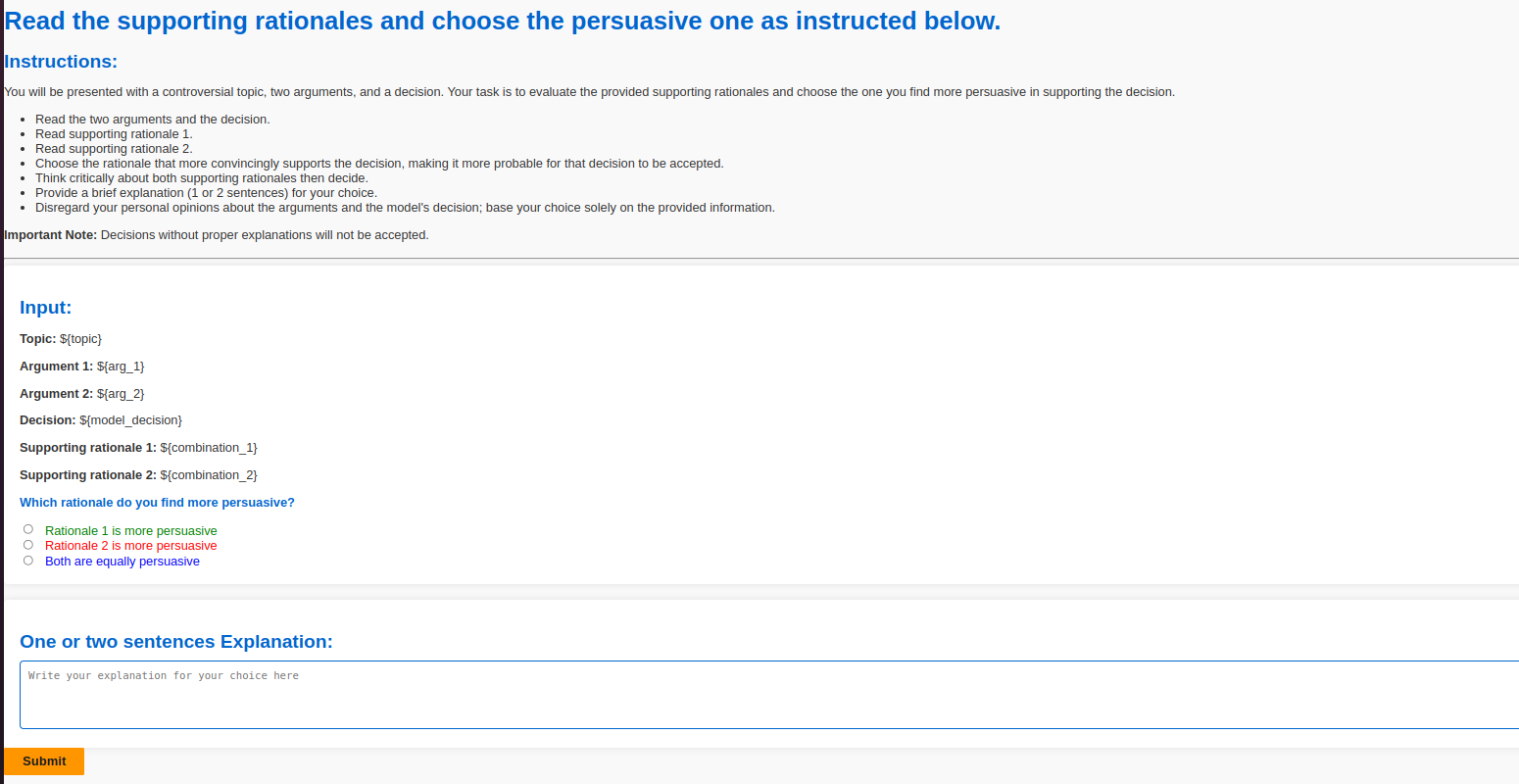}
    \centering
    \caption{\centering 
  A screenshot form  MTurk HIT for persuasion evaluation.
    \label{fig:persuasion-eval}}
\end{figure*}

Figure \ref{fig:persuasion-eval} shows the Mechanical Turk HIT for evaluating pairwise persuasiveness. Workers are prompted to choose between rationale 1, and rationale 2, or indicate that both are equally persuasive. Additionally, they are requested to provide 1-2 sentences as explanations for their decisions.

\section{Persuasion Evaluation with GPT4}
\label{app:persuasion_eval_gpt4}

Table \ref{tab:gpt4_persuasion} shows the components of the prompt we have used in pairwise persuasion ranking of the rationale. 
\begin{table*}
\small
\begin{center}
  \begin{tabular}{|>{\centering\arraybackslash}p{0.9\textwidth}|}
    \hline
    \textbf{GPT4 pairwise persuasion ranking  Prompt} \\
    \hline
    \textbf{Task definition}
  You will be presented with a topic and two arguments, labeled as "ARG1" and "ARG2." One of these arguments, either "ARG1" or "ARG2," is identified as the winner argument ("WINNER\_ARG"). Additionally, two different rationales supporting the winner argument are provided, each indicated by a numerical identifier [1] or [2]. Your task is to \textit{determine which rationale is more persuasive or if they are equally persuasive in supporting the "WINNER\_ARG".} \\
    \hline \\
    \textbf{Formatting Examples}\\





    



\begin{lstlisting}[language=json]
{
//Three formatting examples for each type of output. 
// Actual formatting examples are truncated to save pace. 
}
    
\end{lstlisting}

    \vspace{0.5cm}

    \textbf{Annotation Example} \\
    \begin{lstlisting}[language=json]
     {
     //Actual input 
     }
     Think step by step then decide.
 \end{lstlisting} \\ \hline

  \end{tabular}
\end{center}
\caption{\centering GPT4 based persuasion ranking prompt.  \label{tab:gpt4_persuasion}}
\end{table*}

\section{Characteristics of the generated rationale per model}
\label{app:characteristics}

\begin{figure*}[ht]
    \centering
    \includegraphics[width=\textwidth]{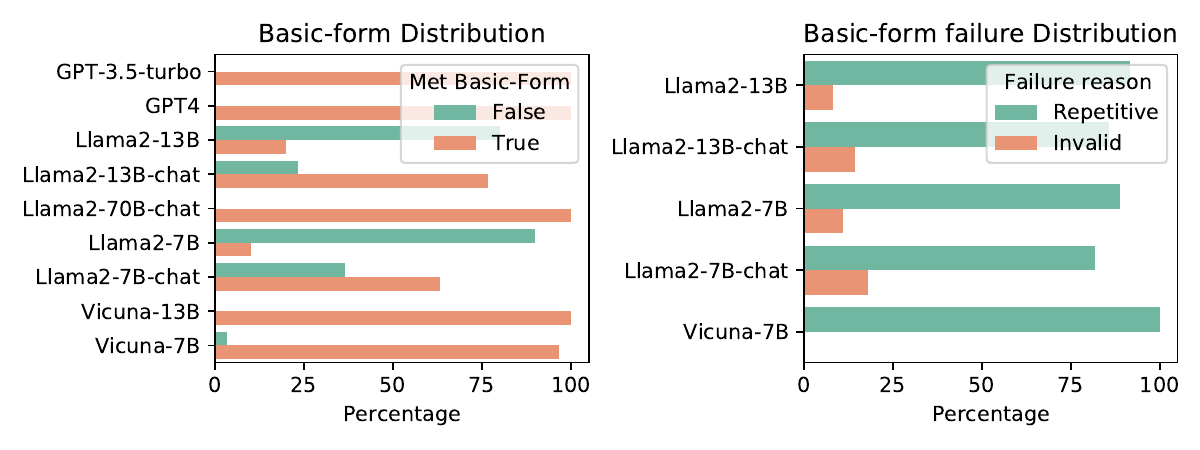}
    \centering
    \caption{\centering 
    Basic-form distribution across models
    \label{fig:basic_form}}
\end{figure*}

\paragraph{Basic Form}
Figure \ref{fig:basic_form} illustrates the percentage of rationales that failed to meet the basic form criteria across all models, along with the breakdown of reasons for failure between invalidity and repetition. The figure shows that Llama-2-7B and Llama-2-13B Chat predominantly generated invalid rationales, suggesting flaws in their reasoning capabilities regarding their choices. Conversely, models of similar sizes that underwent instruction tuning, namely Llama2-7B Chat, Llama2-13B Chat, Vicuna-7B, and Vicuna-13B, demonstrated proficiency in generating meaningful rationales. This emphasizes the significance of instruction tuning in rationalization. Notably, the common observation among samples failing to meet basic requirements was repetition, indicating a tendency among models to reiterate their chosen arguments partially or fully.

\begin{figure*}[ht]
    \centering
    \includegraphics[width=\textwidth]{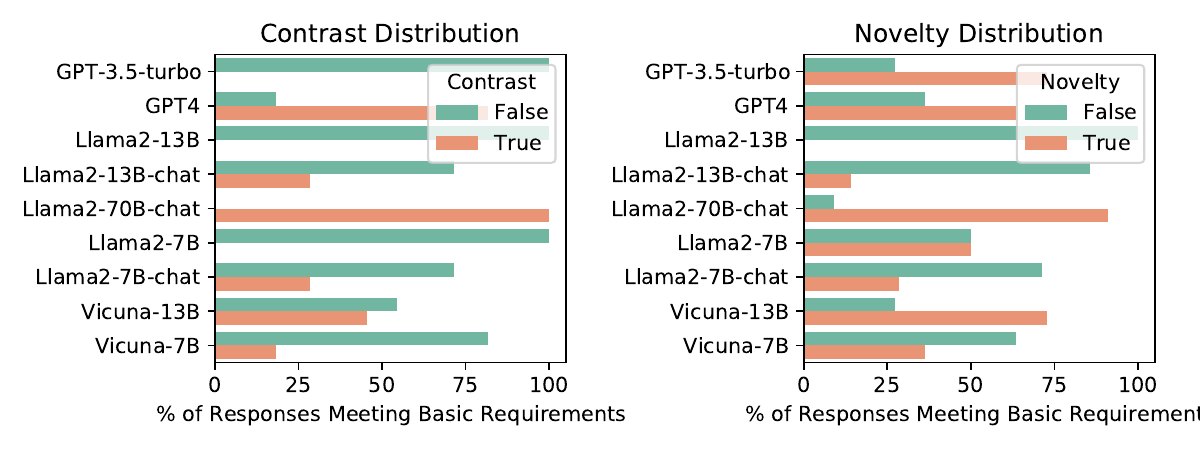}
    \centering
    \caption{\centering 
   Contrast and Novelty distribution among models for samples met basic-form requirements.
    \label{fig:informativeness}}
\end{figure*}

\paragraph{Content Evaluation}
Figure \ref{fig:informativeness} reveals that, among all models, Llama2-70B Chat consistently provided rationales that justified not choosing the alternative argument (contrast). Similarly, GPT4 predominantly generated rationales characterized by contrast. However, the majority of rationales generated by other models did not offer justifications for not selecting the alternative argument.  

In analyzing novelty, it appears that the model scale, demonstrated by Llama2-70B, GPT4, and GPT-3.5-turbo, plays a role in enhancing the models' capacity to offer novel information in their generated rationales, beyond what is explicitly stated in the arguments.

\section{Characteristics of Models Capable of Generating Highly Persuasive Rationales}
\label{app:model_persuasiveness_char}

\textbf{$(1)$ Instruction Tuning:} Among the models we analyzed, those that had not undergone instruction tuning (Llama2-7B and Llama2-13B) failed to provide valid rationales justifying the models' choices. This indicates that mere auto-regressive training is insufficient and that instruction tuning is essential for creating effective rationales.
\textbf{$(2)$ Scale:} The results also highlight that scaling up the parameters within the same model framework enhances persuasiveness. For example, Llama2-70B-chat was found to be more persuasive than its lower parameter counterparts, Llama2-13B-chat and Llama2-7B-chat.
\textbf{$(3)$ Further Tuning with Instructions Obtained from a Stronger LLM:} Vicuna models ranked higher compared to their Llama2 counterparts in the case of the IBM-ArgQ-9.1kPairs dataset, while Vicuna-13B consistently ranked higher on average compared to Llama2-7B-chat and Llama2-13B-chat in terms of the IBM-30k-rank dataset. This suggests that further instruction tuning, based on more advanced models, can improve a model’s capability to generate more compelling rationales.

\section{Rationale Persuasiveness Improvement}
\label{app:persuasion_improved}

\paragraph{Re-prompt the LLM} Table \ref{tab:llm_controlled_persuasion} displays the prompt used to instruct LLMs to generate a more persuasive rationale. The model was prompted to compose $2$ sentences supporting the chosen argument and $2$ sentences indicating reasons for not choosing the alternative argument. This approach ensures that the model includes \textit{contrast} and sufficient detail in its rationalization, which has been shown to enhance persuasiveness.

    \begin{table*}
\small
\begin{center}
  \begin{tabular}{|>{\centering\arraybackslash}p{0.9\textwidth}|}
    \hline
    \textbf{GPT4 pairwise persuasion ranking  Prompt} \\
    \hline
    \textbf{Task definition}
You will receive two arguments, each identified by a numerical identifier [] and a Topic.
Disregarding your own opinion on the topic, given the two arguments, decide which argument you would recommend.
\textbf{Provide a compelling reasoning consists of 2 sentences justifying the argument you choose and 2 sentences stating your reasoning for not choosing the other argument.}
Choose [1] if you if you recommend argument [1] over argument [2].
Choose [2] if you recommend argument [2] over argument [1].
Format your output in a JSON format with "decision" and "reasoning" keys.:  \\
    \hline \\
    \textbf{Formatting Examples}\\





    



\begin{lstlisting}[language=json]
{
//Similar Examples to Table 4
// Actual formatting examples are truncated to save pace. 
}
    
\end{lstlisting}

    \vspace{0.5cm}

    \textbf{Annotation Example} \\
    \begin{lstlisting}[language=json]
     {
     //Actual input 
     }
 \end{lstlisting} \\ \hline

  \end{tabular}
\end{center}
\caption{\centering Pairwise argument ranking with prompting LLM for a more persuasive rationalization.  \textbf{Bolded} part are intended to prompt LLMs to generate contrastive rationales with enough details. \label{tab:llm_controlled_persuasion}}
\end{table*}

\paragraph{Evaluate and Refine}
Table \ref{tab:llm_self_refined_persuasion} shows the prompt used in the \textit{evaluate and refine} method to let the LLM decide if it needs to improve its rationale persuasiveness or not.

   \begin{table*}
\small
\begin{center}
  \begin{tabular}{|>{\centering\arraybackslash}p{0.9\textwidth}|}
    \hline
    \textbf{Evaluate and Refine Prompt} \\
    \hline
    \textbf{Task definition}
You will receive two arguments, each identified by a numerical identifier [] and a Topic.
You will also receive a decission about which argument is better in the format of ARG1 or ARG2 in the format of "WINNER ARG".
You will also receive a "supporting rationale" supporting the "WINNER ARG".
Your task is to decide whether the supporting rationale is "convincing" in terms of supporting the "WINNER ARG".
Choose "YES" if you think the rationale is convincing in supporting the "WINNER ARG".
Choose "NO" if you think it's not convincing in supporting the "WINNER ARG". 
If you choose "YES", generate "no further improvement needed" as your "improved rationale" .
If you choose "NO" generate a more persuasive version of the rationale. \\
    \hline \\
    \textbf{Formatting Examples}\\





    



\begin{lstlisting}[language=json]
{
//2 formatting examples to ensure the model follows them
// Actual formatting examples are truncated to save pace. 
}
    
\end{lstlisting}

    \vspace{0.5cm}

    \textbf{Annotation Example} \\
    \begin{lstlisting}[language=json]
     {
     //Actual input 
     }
 \end{lstlisting} \\ \hline

  \end{tabular}
\end{center}
\caption{\centering Prompt used to instruct the LLM to decide first if it needs to improve the rationale persuasiveness , then improve its persuasiveness if necessary. \label{tab:llm_self_refined_persuasion}}
\end{table*}

\paragraph{Examples of the improved rationales}

Table \ref{tab:argument_summary} presents examples from various improvement methods. Notably, the evaluate and refine method indicated that the original rationale was sufficiently persuasive, suggesting a possible tendency of models to concur with the persuasiveness of their prior outputs. Conversely, when the LLM is explicitly instructed on how to construct a persuasive rationale, the content changes accordingly.

\begin{table*}[htbp]
\centering
\small
\begin{tabular}{|p{14cm}|}
\hline 
  \textbf{Inputs} \\ \hline 
 \textbf{Topic:}  We should promote autonomous cars \\ 
 \textbf{Argument 1:}  Research into the development of fully autonomous vehicles will invariably produce breakthroughs in sensor technology, machine vision, and other related domains that will find application in other fields benefitting society. \\
 \textbf{Argument 2:}  Autonomous vehicles will be safer especially trucks. People who say trucking is the most popular job in America are wrong and only a relatively few people will lose jobs. See \url{https://www.marketwatch.com/story/no-truck-driver-isnt-the-most-common-job-in-your-state-2015-02-12} \\ 
 \textbf{Decision:} Argument 1 \\ \hline
  \textbf{ Generated Rationale} \\ \hline
 \textbf{Llama2-7B-chat:}  The argument that research into the development of fully autonomous vehicles will produce breakthroughs in related domains is more convincing because it highlights the potential for positive impact across multiple fields, rather than simply focusing on the safety benefits of autonomous vehicles. \\ \hline
 \textbf{Llama2-7B-chat-persuasion-prompted:}  While argument 2 mentions potential job losses in the trucking industry, the benefits of promoting autonomous cars far outweigh these potential drawbacks. The development of fully autonomous vehicles will lead to breakthroughs in sensor technology, machine vision, and other related domains, which will have a broader impact on society. Additionally, autonomous vehicles will be safer, including trucks, which will save lives and reduce accidents. Overall, argument 1 presents a more compelling case for promoting autonomous cars. \\ \hline
 \textbf{Llama2-7B-chat-second-stage}
The argument that research into the development of fully autonomous vehicles will produce breakthroughs in related domains is more convincing because it highlights the potential for positive impact across multiple fields, rather than simply focusing on the safety benefits of autonomous vehicles.\\ \hline
\end{tabular}
\caption{\centering Different rationales outputs by different persuasion improvement  methods.}
\label{tab:argument_summary}
\end{table*}

\end{document}